\documentclass[10pt,twocolumn,letterpaper]{article}

\usepackage[pagenumbers]{cvpr} % To force page numbers, e.g. for an arXiv version

\usepackage{graphicx}
\usepackage{amsmath}
\usepackage{amssymb}
\usepackage{booktabs}

\usepackage{arydshln}
\usepackage{multirow}
\usepackage[accsupp]{axessibility}  % Improves PDF readability for those with disabilities.
\usepackage{algorithm}
\usepackage{algorithmic}
\DeclareFontFamily{U}{mathb}{}
\DeclareFontShape{U}{mathb}{m}{n}{
  <-5.5> mathb5
  <5.5-6.5> mathb6
  <6.5-7.5> mathb7
  <7.5-8.5> mathb8
  <8.5-9.5> mathb9
  <9.5-11.5> mathb10
  <11.5-> mathb12
}{}
\DeclareSymbolFont{mathb}{U}{mathb}{m}{n}
\DeclareMathSymbol{\drsh}{3}{mathb}{"EB}
\newcommand{\cornerarrow}{\raisebox{1pt}{$\drsh\ $}}

\usepackage[pagebackref,breaklinks,colorlinks]{hyperref}

\usepackage[capitalize]{cleveref}
\crefname{section}{Sec.}{Secs.}
\Crefname{section}{Section}{Sections}
\Crefname{table}{Table}{Tables}
\crefname{table}{Tab.}{Tabs.}

\def\paperID{5259} % *** Enter the Paper ID here
\def\confName{CVPR}
\def\confYear{2024}

\begin{document}

\newcommand\correspondingauthor{\thanks{Corresponding author.}}

\title{HCPM: Hierarchical Candidates Pruning for Efficient Detector-Free Matching}
% \and
\author{Ying Chen$^{1}$ \quad
        Yong Liu$^{1}$ \quad
        Kai Wu$^{1}$ \quad
        Qiang Nie$^{1}$ \quad
        Shang Xu$^{1}$ \quad
        Huifang Ma$^{2}$ \quad \\
        Bing Wang$^{3}$\footnotemark[2] \quad
        Chengjie Wang$^{1}$\footnotemark[2] \\
        $^{1}$Tencent YouTu Lab \quad $^{2}$ Duke University \quad $^{2}$ The Hong Kong Polytechnic University}
        % {\tt\small \{mumuychen, choasliu, qiangnie, shangxu, jasoncjwang\}@tencent.com}}

% \maketitle
% \renewcommand{\thefootnote}{\fnsymbol{footnote}}
% \footnotetext[1]{These authors contributed equally.}
% \footnotetext[2]{Corresponding author.}
% \footnote{$^{*}$ These authors contributed equally.\\
%           $^{*}$ These authors contributed equally. }

\twocolumn[{%
\renewcommand\twocolumn[1][]{#1}%
\maketitle
\begin{center}
	\includegraphics[width=0.8\linewidth]{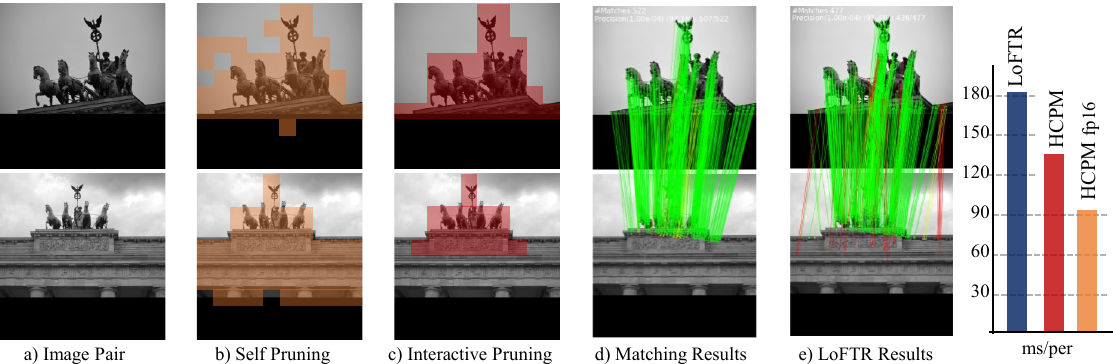}
	\captionsetup{type=figure}\caption{\textbf{Comparison of HCPM with LoFTR~\cite{sun2021loftr}}. The hierarchical pruning stages in our method consist of two pruning processes: self-pruning, which individually prunes candidates, and interactive-pruning, which utilizes interacted information to determine the relative candidates stage-by-stage within the transformer architecture. Our method retains the same accuracy as LoFTR~\cite{sun2021loftr} while reducing the inference time by approximately \textbf{25\%}. Furthermore, by employing FP16 precision, we achieve a decrease in inference time of up to \textbf{50\%}.}
	\label{fig:motivation}
\end{center}
}]
\renewcommand{\thefootnote}{\fnsymbol{footnote}}
% \footnotetext[1]{These authors contributed equally.}
\footnotetext[2]{Corresponding author.}
% \footnote{$^{*}$ These authors contributed equally.\\
%           $^{*}$ These authors contributed equally. }

%%%%%%%%% ABSTRACT
\begin{abstract}
Deep learning-based image matching methods play a crucial role in computer vision, yet they often suffer from substantial computational demands. To tackle this challenge, we present HCPM, an efficient and detector-free local feature-matching method that employs hierarchical pruning to optimize the matching pipeline. In contrast to recent detector-free methods that depend on an exhaustive set of coarse-level candidates for matching, HCPM selectively concentrates on a concise subset of informative candidates, resulting in fewer computational candidates and enhanced matching efficiency. The method comprises a self-pruning stage for selecting reliable candidates and an interactive-pruning stage that identifies correlated patches at the coarse level. Our results reveal that HCPM significantly surpasses existing methods in terms of speed while maintaining high accuracy. The source code will be made available upon publication.
\end{abstract}
\vspace{-5mm}

\section{Introduction}
\label{sec:intro}

Local feature matching is a fundamental task and serves as the foundation for various 3D computer vision applications, such as Structure from Motion (SfM), autonomous driving, and visual localization. This field has achieved significant improvements in matching accuracy through both detector-based methods, such as Superglue~\cite{sarlin2020superglue}, and detector-free methods like LoFTR~\cite{sun2021loftr}. However, with the revolution of transformer architecture, both methods achieve high accuracy at the expense of increased computational complexity.
This trade-off between accuracy and efficiency in local feature matching constrains real-time performance, and exacerbates power consumption.

To enhance feature-matching efficiency, recent studies have primarily focused on keypoint selection for detector-based methods. Works such as \cite{shi2022clustergnn} and \cite{cai2023htmatch} propose restricting matching within a cluster of selected keypoints based on GNN. Similarly, LightGlue~\cite{lindenberger2023lightglue} aims to prune keypoints with low matchable confidence in the early stages. However, these methods rely on feature detectors from detector-based methods and have only proven useful with sparse inputs. In contrast, detector-free techniques are known for their robustness with pixel-wise dense matches compared to using a predefined keypoint detector. Leveraging the transformer's ability to capture long-distance dependencies, detector-free methods like LoFTR~\cite{sun2021loftr} have achieved significant performance gains through global feature aggregation and pixel-level feature refinements. However, since computation and memory costs increase quadratically within the transformer architecture, detector-free methods face substantial computational challenges.

% 引入token pruning等在segmentation/detection等transformer架构的应用 
Being the first to explore acceleration for detector-free methods, we intend to keep the dense matching advantages and reduce the complexity on the semantic level within the transformer architecture by token pruning.
Significant contributions in token pruning, such as \cite{chen2023sparsevit, fang2023unleashing, fayyaz2022adaptive, tang2023dynamic, tang2022patch, wei2023joint}, have innovatively employed pruning strategies to eliminate redundant inputs with minimal informative value. 
This technique has proven instrumental in segmentation and detection tasks, while the application in image matching tasks remains unexplored. 
Directly applying token pruning to image matching usually results in substantial performance degradation. 
This is primarily due to the fact that the redundancy reduction of token pruning is limited to a single image, where the co-visible area is crucial for matching algorithms \cite{sun2021loftr, chen2022guide}.

%  我们受此启发，也对image matching的候选点进行重要性计算和筛选，来进行加速，具体描述实现。总结贡献点。
Building on the concept of token pruning, we introduce HCPM, a method that hierarchically prunes unnecessary candidates while retaining the dense benefits of detector-free methods. HCPM is designed to emulate human behavior, prioritizing visually significant features such as static buildings and signposts within the co-visible area, which are generally more crucial for local feature matching than transient natural elements like trees or the sky. Our method employs a hierarchical pruning process to select matching candidates. In the initial stage, self-pruning is used to identify the top-k candidates based on a confidence score generated from a straightforward yet effective activation mechanism, where k is determined by a hyper-parameter ratio $\alpha$. The selected candidates then proceed to an interactive-pruning phase, which gradually aggregates information and eliminates unrelated candidates. This phase is characterized by multiple self-cross attention modules that facilitate the extraction and integration of co-relative features through a cross-attention mechanism. Additionally, drawing inspiration from OETR~\cite{chen2022guide}, we use the co-visible area to supervise our differentiable selection process, with co-visible are supervision at each iteration directing the network's focus on the co-visible candidates. To fully automate the selection process, we propose Gumbel-Softmax~\cite{jang2016categorical} learned masks following each self-cross attention layer. This approach refines the final set of candidates without the need for manually set thresholds, enhancing the overall efficiency and effectiveness of the method.

To summarize, we aim to provide several critical insights of efficient local feature matching:
\begin{itemize}
\item We introduce HCPM, an efficient detector-free matching approach that employs a self-pruning and interactive-pruning to reduce matching redundancy and disturbances.
%\vspace{-2mm}
\item Our method provides a differentiable selection strategy, leveraging co-visible information to supervise the selection process.
%\vspace{-2mm}
\item Our experiments reveal that HCPM attains competitive performance with nearly 50\% reduced computational cost, closely approaching state-of-the-art methods in numerous vision tasks.
\end{itemize}

\begin{figure*}[ht]
    \centering
    \includegraphics[width=0.95\textwidth]{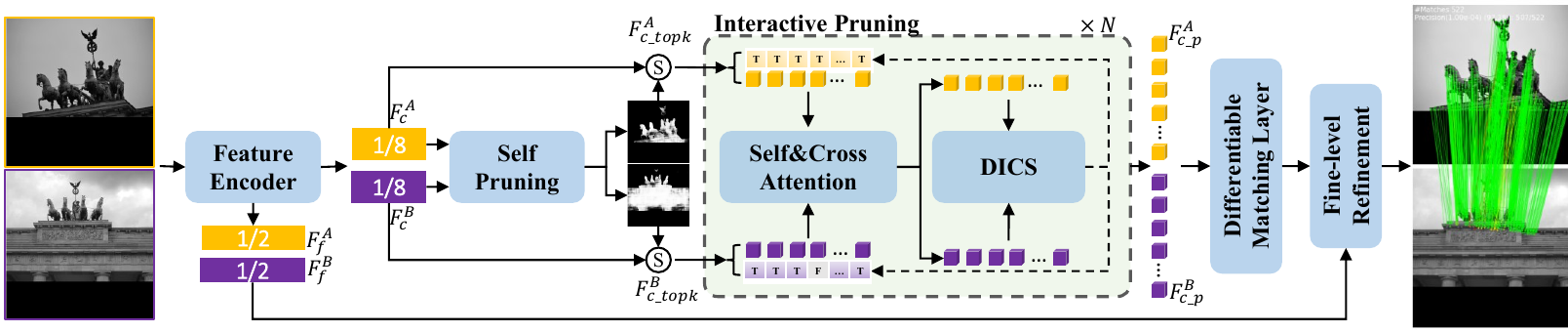}
    \caption{\textbf{Architecture of HCPM.} Upon obtaining coarse feature maps $F_c^A, F_c^B$ from the feature encoder module, they are fed into a self-pruning module for a static ratio top-k selection, denoting the selected feature as $F_{c\_topk}^A, F_{c\_topk}^B$. Subsequently, $F_{c\_topk}^A, F_{c\_topk}^B$ and masks are input into interactive-pruning blocks, which encompass a self-cross attention and a Differentiable Interactive Candidates Selection (DICS) module. The self-cross attention enhances the feature, which then undergoes an automated selection process via the DICS, resulting in pruned candidates. After $N_c$ times iteration, we obtain pruned candidate features $F_{c\_p}^A$ and $F_{c\_p}^B$. Ultimately, the pruned coarse-level features generate a matching matrix and collaborate with fine-level features to regress matching positions, as in LoFTR~\cite{sun2021loftr}.}
    \label{fig:method}
\end{figure*}

\section{Related works}

\subsection{Local Feature Matching}
Local feature matching is a fundamental research area in computer vision, encompassing a long and complex pipeline that includes detectors, descriptors, and matchers. Recently, these approaches can be classified into detector-based and detector-free methods. Detector-based methods~\cite{sarlin2020superglue, shi2022clustergnn, chen2021learning, lindenberger2023lightglue}, which concentrate on sparse keypoints matching, typically offer computational efficiency but may lack robustness. In contrast, detector-free methods~\cite{sun2021loftr, chen2022aspanformer, savinov2017quad, huang2022adaptive}, which consider whole image coarse-level pixel as a potential feature point, provide greater robustness at the expense of increased computational demands. Their success can be largely attributed to the integration with Transformers~\cite{vaswani2017attention}, which effectively capture long-range correlations. Nonetheless, the considerable computational complexity of Transformers presents a significant challenge for efficiency.

Although some detector-based methods, such as ClusterGNN~\cite{shi2022clustergnn} and LightGlue~\cite{lindenberger2023lightglue}, have attempted to accelerate matching, their performance remains inferior to that of detector-free methods. ClusterGNN~\cite{shi2022clustergnn} employs a clustering approach for learning the feature matching task, adaptively partitioning keypoints into distinct subgraphs to minimize redundant connectivity and utilizing a coarse-to-fine paradigm to mitigate misclassification within images. Building upon SuperGlue~\cite{sarlin2020superglue}, LightGlue~\cite{lindenberger2023lightglue} demonstrates adaptability to problem difficulty, facilitating faster inference on image pairs that are intuitively easier to match due to larger visual overlap or limited appearance change. However, in comparison with detector-free methods~\cite{sun2021loftr, chen2022aspanformer, tang2022quadtree, huang2022adaptive}, detector-based methods still exhibit a noticeable gap in precision. While detector-free methods encompass all coarse-level candidates for matching and are generally slower due to the processing of a large number of candidates, HCPM leverages only a subset of the dense correspondences, achieving a balance between accuracy and efficiency.

\subsection{Efficient Transformer}
The computational burden of the Transformer is dictated by the number of tokens and the intricacy of its architecture. Consequently, we classify efficient transformers into two categories: those focusing on token pruning~\cite{chen2023sparsevit, fang2023unleashing, fayyaz2022adaptive, tang2023dynamic, tang2022patch, wei2023joint}, and those concentrating on architectural design~\cite{lineartransformer, lee2022composite, kitaev2020reformer, beltagy2020longformer, shen2021efficient, dao2022flashattention}.

In the first category, TPS~\cite{tang2022patch} advocates a top-down strategy to eliminate redundant patches in vision transformers, while DTP~\cite{tang2023dynamic} presents an early exit of tokens for semantic segmentation. SparseViT~\cite{chen2023sparsevit} reexamines activation sparsity for window-based vision transformers, achieving a 50\% latency reduction with 60\% sparsity. Despite the success of token pruning, the computational expense of vanilla attention at high resolution remains daunting. This has prompted the development of architectural approximations as highlighted in~\cite{lineartransformer, choromanski2020rethinking, tang2022quadtree, wang2020linformer, zaheer2020big}. In the realm of image matching, vanilla attention and Linear Attention~\cite{lineartransformer} are two notable attention architectures. ~\cite{tang2022quadtree} employs QuadTree Attention and a tree data structure with four children per internal node to reduce computational complexity from quadratic to linear, selecting the top K patches with the highest attention scores to focus on relevant regions in the subsequent level. ASpanFormer~\cite{chen2022aspanformer} introduces an attention mechanism that adjusts the attention span based on the computed flow maps and the adaptive sampling grid size. Although their performance exceeds that of linear attention, they still confront significant computational challenges.

While token pruning proves effective in numerous vision tasks, it has not been applied in detector-free matching. Most detector-free matching efforts concentrate on transformer architecture design and improving accuracy at the expense of time consumption. Inspired by token pruning, our approach circumvents complex architectural designs and employs hierarchical pruning stages to boost efficiency while preserving accuracy.

\section{Methods}
We first provide an introduction to Hierarchical Candidates Pruning for Efficient Detector-Free Matching (HCPM) in Sec.~\ref{3.1}. Then, we describe our self-pruning strategy in Sec.~\ref{3.2}. In Sec.~\ref{3.3}, we discuss interactive pruning in detail, which primarily consists of two elements: Differentiable Interactive Candidate Selection (DICS) and Interactive-Pruning Attention (IPA). Finally, we introduce the supervision of the training pipeline in Sec.~\ref{3.4}.

% %\vspace{-1mm}
\subsection{Preliminary and Overview}\label{3.1}
\noindent\textbf{Preliminary}.
We briefly review LoFTR~\cite{sun2021loftr}, which applies a coarse-to-fine approach to produce dense matches with higher accuracy. It uses a ResNetFPN feature encoder to extract coarse-level feature maps $F_c^A, F_c^B$ at $\frac{1}{8}$ of the original image spatial dimension and fine-level feature maps $F_f^A, F_f^B$ at $\frac{1}{2}$ of the original image spatial dimension for each image. The coarse features $F_c^A, F_c^B$ are then fed into a coarse-matching module, updated with a linear Transformer-based self-cross (SC) attention module. Subsequently, these coarse-level features are leveraged to learn a matching confidence matrix by a Differentiable Matching Layer, yielding coarse matching predictions. Finally, a Fine-level Refinement module predicts the sub-pixel coordinates for each coarse-level center feature using the previously obtained coarse-level matching candidates.

\noindent\textbf{Overview.}
From the aforementioned process, it is evident that all coarse features contribute equally within the Self-Cross (SC) attention module. Our Hierarchical Coarse-to-fine Pruning Module (HCPM) utilizes a hierarchical pruning technique to streamline candidate selection, thereby enhancing both efficiency and effectiveness by selecting informative matching candidates in a hierarchical manner. As depicted in Figure~\ref{fig:method}, the proposed HCPM adheres to the principal coarse-to-fine approach in LoFTR, primarily comprising a hierarchical pruning strategy combined with self-pruning and interactive-pruning.

\subsection{Self Pruning}\label{3.2}
Traditional methods of feature and descriptor extraction, such as SIFT~\cite{Lowe2004sift}, typically identify keypoints in areas with significant gradients. With the advent of deep learning, a variety of deep learning-based keypoint and descriptor methods have emerged, including D2Net~\cite{D2-Net}, R2D2~\cite{R2D2}, and Superpoint~\cite{Detone2018superpoint}. Most of these methods select local response feature maxima as keypoints, with the majority of keypoints used for matching located on rigid, static objects. Conversely, areas like the sky, pedestrians, and plants are generally unsuitable for keypoints and pose challenges for matching, as most of them are not informative.

In this study, we contend that not all pixels in images are equally important for matching; some pixels are informative for matching while others introduce noise and disturbance. Therefore, in the detector-free matching process, we should not involve all coarse feature candidates in the matching procedure. To address this, we propose a self-pruning method that autonomously filters out non-informative candidates for matching, thereby concentrating on the most informative candidates for enhanced efficiency.

In HCPM, the feature encoder utilizes a ResNet-FPN architecture for feature extraction, which effectively captures multi-level fusion information with abundant semantic content. The coarse-level features are denoted as $F_c^A, F_c^B$. To process these features, we employ a simple MLP to encode the features from their original 256 channels to a single channel representation. Subsequently, we apply a Sigmoid function to transform the encoded features into a candidate informative score:

\begin{equation}
%\vspace{-1mm}
\begin{aligned} 
S_{A,B}=\operatorname{Sigmoid}(\mathbf{MLP}(F_c^{A,B})) \in [0,1]\\
\label{self-pruning}
\end{aligned}
% \vspace{-3mm}
\end{equation}

Given an informative selection ratio $\alpha$, we can calculate the selected number from the input pixel dimension, as the selected $k$ number is: $k=\frac{H}{8}\times \frac{W}{8} \times \alpha$. We first gather candidates with the highest importance scores, $F_{c\_topk}^i=F_c^i(\operatorname{TopK}(S_i)), i\in\{A, B\}$. Only selected candidates $F_{c\_topk}^A, F_{c\_topk}^B$ will participate in the next stage of the matching pipeline. This method aims to eliminate unsuitable candidates while preserving informative regions, striking a balance between efficiency and accuracy. As informative candidates in the matching process can always reconstruct depth from multiview triangulation, we use depth signals for self-pruning results supervision. Since self-pruning is the initial step in the process, it is crucial not to filter out informative areas excessively. Therefore, some redundancy is retained for less informative regions, ensuring that the subsequent stages of the matching process have sufficient information to work with; here, we set $\alpha$ to 0.5. Although this process is much simpler than other learning-based measures, it introduces smaller computational overhead and proves to be quite effective in practice.

\subsection{Interactive-Pruning}\label{3.3}
Unlike the self-pruning method that only involves single-image information, our method is based on a detector-free framework, which typically takes two images as input. This approach allows for more comprehensive interaction and utilization of information between the two images. In contrast to self-pruning with a top-k selection, we design a differentiable interactive candidate selection module (DICS) to select candidates after one self-cross attention process, enabling an automatic selection process without any hyper-parameters. Finally, we discuss two interactive pruning attention methods after DICS, such as direct-pruning, which discards pruned features, or retaining them as an implicit-pruning method.

\begin{figure}[htp]
    \centering
    \includegraphics[width=0.45\textwidth]{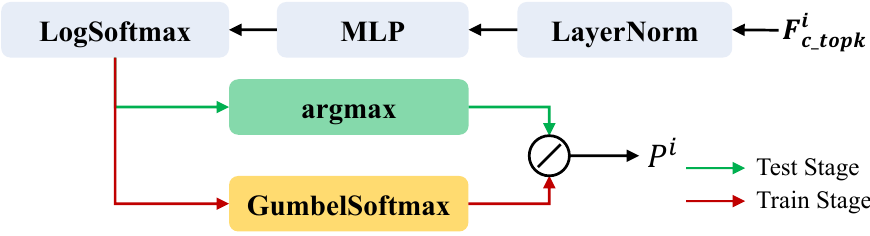}
    \caption{\textbf{Differentiable interactive candidate selection(DICS).} }
    \label{fig:DICS}
\end{figure}

\subsubsection{Differentiable Interactive Candidate Selection}
In the self-pruning process, we utilize depth information to supervise the self-pruning candidates, enabling a differentiable calculation as we compute a sigmoid confidence map for supervision. Then, in the interactive pruning, the utilization of information from both images becomes crucial. In previous works, OETR~\cite{chen2021learning} calculates the co-visible area between two images as a preprocessing module in the matching pipeline, and Adamatcher~\cite{huang2022adaptive} estimates a co-visible area after $N_c$ times self-cross attention, both demonstrating that constraining candidates to the co-visible area can improve matching performance. Inspired by this, we designed a Differentiable Interactive Candidate Selection (DICS) module to calculate the co-visible candidates after self-pruning. Unlike previous research, which uses a semantic or a detection pipeline to calculate the co-visible area, we employ a differentiable selection method.

After the self-pruning process, the residual candidate features $F_{c\_topk}^A, F_{c\_topk}^B$ are fed into multiple self-cross attention layers to enhance feature performance:
\begin{equation}
\begin{aligned} 
F_{c\_topk}^{A,B} = SA(F_{c\_topk}^{A,B}, F_{c\_topk}^{A,B}, M^{A,B}, M^{A,B}) \\
F_{c\_topk}^{A,B} = CA(F_{c\_topk}^{A,B}, F_{c\_topk}^{B,A}, M^{A,B}, M^{B,A}) \\
\label{eq:SCA}
\end{aligned}
\end{equation}
Here, $SA$ denotes the self-attention operation, and $CA$ represents the cross-attention operation, $M^{A,B}$ is the mask of top-k candidates. The differentiable interactive candidate selection process aims to efficiently select and utilize the most relevant and reliable candidates after self-cross attention for improved performance in the matching process.

Unlike conventional attention mechanisms that use a padding mask or no mask, our approach employs the DICS module, which calculates which candidates need to be pruned after every self-cross attention process. If a token requires pruning, the corresponding mask is changed to False. To determine which candidates to keep or prune after each self-cross attention operation, we introduce a lightweight sub-network to predict the keep probability $p^{A,B} \in \mathbb{R}^{N, 2}$ for candidates $F^{A,B}_{c\_topk}$.

\begin{equation}
\begin{aligned} 
p^{A,B}=Softmax(MLP(Norm(F^{A,B}_{c\_topk}))) \\
\end{aligned}
\end{equation}

After predicting the probability $p^{A,B}$, we need to set a ratio to decide the top-k probability candidates to keep. However, different pairs are associated with different co-visible areas, making the ratios of selected candidates challenging to design. To address this issue, we relax the sampling of discrete top-K masks to a continuous approximation, using the Gumbel-Softmax~\cite{jang2016categorical, maddison2016concrete} to render the discrete decision differentiable:

\begin{equation}
\begin{aligned} 
&P^{A,B}=1-GumbelSoftmax(p^{A,B})[:,0] \in \{0, 1\}^N  \\
&M^{A,B}=P^{A,B}\times M^{A,B} \in \{0, 1\}^N \\
\end{aligned}
\end{equation}
More details are shown in Figure~\ref{fig:DICS}. In the training stage, we obtain the two-channel one-hot vectors and select the first channel as $P^{A,B}$ for interactive selection. For the testing stage, we use argmax, with $P^{A,B}=1-argmax(p^{A,B})$ for selection. In contrast to self-pruning, which directly discards pruned candidates, our module retains them but implicitly prunes them using masks. This approach enables a more flexible and efficient pruning process, as pruned candidates can still be accessed and utilized if needed, while the mask ensures they are not considered during the main computation.

\subsubsection{Interactive Pruning Attention}
We review the vanilla and linear attention mechanisms, commonly used in image matching models. Given three inputs: query $Q$, key $K$, and value $V$, vanilla attention computes a weighted sum of the value based on the query-key relationship, with a complexity of $O(n^2)$. Linear attention~\cite{lineartransformer} reduces this complexity by replacing the softmax operator with the product of two kernel functions, where $\phi(\cdot) = \mathrm{elu}(\cdot) + 1$. The above processes can be formulated as:

% \vspace{-3mm}
\begin{equation}
\begin{aligned} 
\operatorname{Att}(\mathbf{Q}, \mathbf{K}, \mathbf{V})=\operatorname{softmax}\left(\mathbf{Q K}^{\top}\right) \mathbf{V} \\
\operatorname{LinAtt}(\mathbf{Q}, \mathbf{K}, \mathbf{V})=\phi(\mathbf{Q}) (\phi(\mathbf{K})^{\top} \mathbf{V}) \\
\label{eq:att}
\end{aligned}
% \vspace{-2mm}
\end{equation}

It should be noted that the above attention paradigm is based on densely fully-connected layers, which require computing the attention map for all query-key pairs. In our work, we develop an interactive pruning attention to replace it, thus avoiding the involvement of irrelevant information during the feature interaction process. As we employ linear attention in our method, after the DICS module, we obtain a one-hot mask for candidates, with zero representing pruning and one for keeping.

\begin{figure}[t]
\centering
\resizebox{.48\textwidth}{!}{
    \begin{tabular}{@{}cc@{}}
        \multicolumn{1}{c}{\includegraphics[]{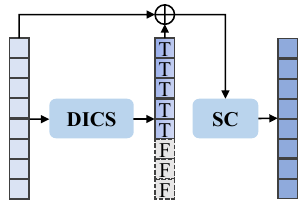}} & 
        \multicolumn{1}{c}{\includegraphics[]{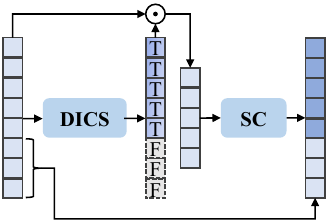}} \\
        \multicolumn{1}{c}{a) Implicit Pruning} & \multicolumn{1}{c}{b) Direct Pruning} 
    \end{tabular}
}
\caption{\textbf{Two sparse pruning modules explored in HCPM.} In implicit-pruning, $\bigoplus$ represents combining the feature and the one-hot mask selected by the DICS module, which are then input into the self-cross (SC) attention module. For direct-pruning, we use $\bigodot$ to input only the selected feature into the self-cross (SC) attention module.}
\label{fig:S-Prune}
\end{figure}

We explored two different interactive-pruning attention methods, as shown in Figure~\ref{fig:S-Prune}. The first one is direct pruning attention, which only involves selected candidates to participate in attention. In this method, we use a selection function, denoted as $\operatorname{Sel}(\cdot)$, to filter out the pruned candidates based on the mask. The direct pruning attention method can be formulated as:

% \vspace{-3mm}
\begin{equation*}
\begin{aligned} 
&\operatorname{DirectAtt}(\mathbf{Q}, \mathbf{K}, \mathbf{V}, \mathbf{M_{Q}}, \mathbf{M_{KV}})\\
&=\operatorname{LinAtt}(\operatorname{Sel}(\mathbf{Q},\mathbf{M_{Q}}), \operatorname{Sel}(\mathbf{K},\mathbf{M_{KV}}), \operatorname{Sel}(\mathbf{V},\mathbf{M_{KV}}))\\
\label{eq:direct}
\end{aligned}
\vspace{-3mm}
\end{equation*}
The second method is an implicit pruning method, which uses a mask to shield the candidates that need to be pruned by element-wise multiplication with the mask. In this method, the pruned candidates are not completely removed, but their influence on the attention process is minimized. The implicit pruning attention method can be formulated as:

\vspace{-3mm}
\begin{equation}
\begin{aligned} 
&\operatorname{ImplicitAtt}(\mathbf{Q}, \mathbf{K}, \mathbf{V}, \mathbf{M_{Q}}, \mathbf{M_{KV}})\\
&=\phi(\mathbf{Q})\mathbf{M_Q}((\phi(\mathbf{K})\mathbf{M_{KV}})^{\top} \mathbf{V}\circ\mathbf{M_{KV}})  \\
\label{eq:implicit}
\end{aligned}
% \vspace{-2mm}
\end{equation}
Both methods aim to efficiently select and utilize the most relevant and reliable keypoints for improved performance in the matching process while reducing the computational complexity of the attention mechanism. More details are shown in Figure~\ref{fig:S-Prune}. As the direct pruning method discards the pruned candidates across multiple stages of transformer blocks, we squeeze the pruned candidates' features to their original position to prevent information loss in $N_c$ times iteration. More experiments are presented in Table~\ref{tab:ablation} and Table~\ref{tab:ablation_interactive}, and we select implicit interactive-pruning as the final choice.

\subsection{Loss and Supervision}\label{3.4}
Our loss function primarily consists of two parts: pruning loss and matching loss.

\noindent\textbf{Pruning Loss.}
The pruning stage comprises both self-pruning and implicit interactive-pruning processes. We treat both as per-candidate binary classification tasks, as OETR~\cite{chen2022guide} and Adamatcher~\cite{huang2022adaptive} use co-visible area for candidates selection supervision. For self-pruning, it is supervised by the depth validity information $D$, denoted as $\widehat{D}_v = D > 0$, where a value of 1 represents depth greater than 0. For implicit interactive-pruning, we use selected co-visible candidates for supervision, which are calculated based on depth, camera poses, intrinsic parameters, and self-pruning results. The self-pruning loss $\mathcal{L}_{SPrune}$ uses cross-entropy (CE) loss. On the other hand, the interactive-pruning loss $\mathcal{L}_{IPrune}$ uses Focal Loss \cite{Lin2017fl}, as interactive results are influenced by self-pruning results, leading to an imbalance in candidate classes regarding co-visibility.

\begin{equation}
\begin{aligned} 
&\mathcal{L}_{SPrune}=CE(S^A_c, \widehat{D}_v^A)+CE(S^B_c, \widehat{D}_v^B)\\
&\mathcal{L}_{IPrune}=FL(M^A_c, \widehat{D}_v^A)+FL(M^B_c, \widehat{D}_v^B)
\end{aligned}
\end{equation}

\noindent\textbf{Matching Loss.}
The coarse matching loss $\mathcal{L}_{c}$ and fine matching loss $\mathcal{L}_{f}$ are the same as in LoFTR ~\cite{sun2021loftr}.

Our final loss is balanced as follows:
\begin{equation}
\mathcal{L} = 0.5*\mathcal{L}_{SPrune} + 0.3*\mathcal{L}_{IPrune} + 1.0*\mathcal{L}_{c} + 1.0*\mathcal{L}_{f}
% \vspace{-3mm}
\end{equation}

\section{Implementation Details}
Following the approach in LoFTR~\cite{sun2021loftr}, we train HCPM on the MegaDepth datasets without applying any data augmentation techniques. The training process utilizes the AdamW optimizer with an initial learning rate of $8 \times 10^{-3}$ and a batch size of 2. The model converges in 1.5 days using 8 V100 GPUs. The image feature extractor is based on a ResNet-FPN~\cite{He2016,Lin2017} architecture, which extends HCPM to resolutions of $\{\frac{1}{8}, \frac{1}{2}\}$. We set the self-pruning ratio $\alpha$ to 0.5 and the patch window size $w$ for refinement to 5. The number of channels for the $F_c^{A,B}$ and $F_f^{A,B}$ features is 256 and 128, respectively. To save GPU memory usage during training, we sample 30 percent of self-pruning selected candidate matches from the match proposals for supervision in the sub-pixel refinement module.

\begin{figure*}[ht]
    \centering
    \includegraphics[width=0.9\textwidth]{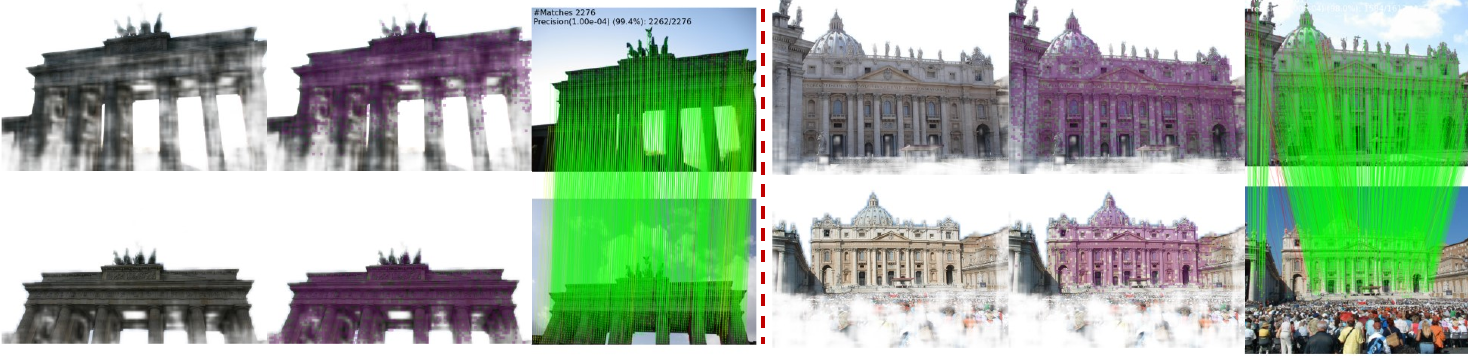}
    \caption{\textbf{Iterative-Pruning and Matching Visualization.} We have chosen two scenes for visualization. The first column shows the self-pruning result, and the second column displays the interactive-pruning result. Since we supervised with the co-visible area, the visualization may appear inconsistent. The third column presents the matching result.}
    \label{fig:method}
\vspace{-2mm}
\end{figure*}

\section{Experiments}
We evaluate our approach against the current state-of-the-art methods in the domains of homography estimation, relative pose estimation, and visual localization. All methods are assessed under identical conditions, encompassing the number of iterations, outlier filtering methodologies, and ratio. Moreover, we scrutinize the influence of our design choices on the comprehensive performance.

\begin{table}[t]
\caption{Evaluation on HPatches~\cite{balntas2017hpatches} for homography estimation.}
\label{tab:HPatches_result}
\resizebox{.45\textwidth}{!}{
\begin{tabular}{lllll}
\toprule
\multicolumn{1}{l}{\multirow{2}{*}{\textbf{Method}}} & \multicolumn{3}{c}{\textbf{Homography est. AUC}}   &\multicolumn{1}{l}{\multirow{2}{*}{\textbf{ms/per}}}     \\\cline{2-4}
            & @3px                   & @5px                   & @10px                                    \\\midrule
SuperGlue~\cite{sarlin2020superglue}~\tiny{CVPR'19}   & \multicolumn{1}{r}{53.9} & \multicolumn{1}{r}{68.3} & \multicolumn{1}{r}{81.7} & \multicolumn{1}{r}{86.6}    \\
SGMNet~\cite{chen2021learning}~\tiny{ICCV'21}      & \multicolumn{1}{r}{54.8} & \multicolumn{1}{r}{68.9} & \multicolumn{1}{r}{82.3} & \multicolumn{1}{r}{184.3}     \\
LightGlue~\cite{lindenberger2023lightglue}~\tiny{ICCV'23}   & \multicolumn{1}{r}{50.6} & \multicolumn{1}{r}{66.3} & \multicolumn{1}{r}{80.9} & \multicolumn{1}{r}{57.4}     \\
\midrule
LoFTR~\cite{sun2021loftr}~\tiny{CVPR'21}       & \multicolumn{1}{r}{65.9} & \multicolumn{1}{r}{75.6} & \multicolumn{1}{r}{84.6} & \multicolumn{1}{r}{116.1} \\
QuadTree~\cite{tang2022quadtree}~\tiny{ICLR'22}   & \multicolumn{1}{r}{66.3} & \multicolumn{1}{r}{76.2} & \multicolumn{1}{r}{84.9}    & \multicolumn{1}{r}{179.2}  \\
AspanFormer~\cite{chen2022aspanformer}~\tiny{ECCV'22} & \multicolumn{1}{r}{\textbf{67.4}} & \multicolumn{1}{r}{\textbf{76.9}} & \multicolumn{1}{r}{\textbf{85.6}} & \multicolumn{1}{r}{147.9} \\
MatchFormer~\cite{wang2022matchformer}~\tiny{ACCV'22} & \multicolumn{1}{r}{63.7} & \multicolumn{1}{r}{73.8} & \multicolumn{1}{r}{83.8} & \multicolumn{1}{r}{176.2} \\
% AdaMatcher~\cite{huang2022adaptive}~\tiny{CVPR'23}        & \multicolumn{1}{r}{67.1} & \multicolumn{1}{r}{77.0} & \multicolumn{1}{r}{85.7} & \multicolumn{1}{r}{213.9} \\
AdaMatcher~\cite{huang2022adaptive}~\tiny{CVPR'23}        & \multicolumn{1}{r}{65.5} & \multicolumn{1}{r}{75.5} & \multicolumn{1}{r}{84.7} & \multicolumn{1}{r}{178.4} \\
HCPM        & \multicolumn{1}{r}{64.5} & \multicolumn{1}{r}{74.2} & \multicolumn{1}{r}{83.7} & \multicolumn{1}{r}{74.2} \\
\cornerarrow FP16 & \multicolumn{1}{r}{64.1} & \multicolumn{1}{r}{74.0} & \multicolumn{1}{r}{83.4} & \multicolumn{1}{r}{\textbf{50.6}} \\\bottomrule  
\end{tabular}}
\end{table}
% PATS~\cite{ni2023pats}~\tiny{CVPR'23}        & \multicolumn{1}{r}{66.7} & \multicolumn{1}{r}{76.8} & \multicolumn{1}{r}{85.9} & \multicolumn{1}{r}{2328.7} \\
% CasMTR-4c~\cite{cao2023improving}~\tiny{ICCV'23}   & \multicolumn{1}{r}{67.5} & \multicolumn{1}{r}{77.1} & \multicolumn{1}{r}{86.3} & \multicolumn{1}{r}{193.2} \\
% AdaMatcher~\cite{huang2022adaptive}~\tiny{CVPR'23}        & \multicolumn{1}{r}{65.5} & \multicolumn{1}{r}{75.5} & \multicolumn{1}{r}{84.7} & \multicolumn{1}{r}{178.4} \\

\subsection{Homography Estimation}
HPatches~\cite{balntas2017}, a well-known image matching evaluation dataset, consists of 116 scenes, including 57 sequences with substantial illumination variations and 59 sequences with notable viewpoint changes, enabling us to evaluate our method under diverse conditions.

\noindent\textbf{Evaluation protocol.} Following \cite{Detone2018superpoint, sarlin2020superglue, zhou2021patch2pix}, we employ corner correctness to assess the performance of the estimated homography. Four corners from the first reference image are warped to the other image using the estimated homography. To ensure a fair comparison with other methods, all images are resized so that their shorter dimensions equal 480. We use OpenCV RANSAC as the robust estimator, in accordance with \cite{zhou2021patch2pix}. As per ~\cite{sun2021loftr}, we report the area under the cumulative curve (AUC) of the corner error up to threshold values of 3, 5, and 10 pixels, respectively. Furthermore, we calculate the average time for pair matches computation. For detector-based methods, which include extractor time, the result is denoted as $ms/per$. All evaluation is performed using an open-source toolbox\footnote{https://github.com/GrumpyZhou/image-matching-toolbox}.

\noindent\textbf{Baseline methods.} We chose several open-source implementations for comparison with our HCPM method, including both typical and recent research approaches. For detector-based methods, we selected SuperGlue~\cite{sarlin2020superglue}, SGMNet~\cite{chen2021learning}, and LightGlue~\cite{lindenberger2023lightglue}, all of which are combined with the SuperPoint~\cite{Detone2018superpoint} extractor. For detector-free methods, we selected LoFTR~\cite{sun2021loftr}, the pioneering work in detector-free methods, as well as QuadTree~\cite{tang2022quadtree} and AspanFormer~\cite{chen2022aspanformer}, which redesigned the LoFTR transformer module. Additionally, we included recent works addressing scale variation, such as AdaMatcher~\cite{huang2022adaptive}, along with other recent research contributions. Besides, as HPatches datasets are with more informative candidates, most of the pixel is static and informative, so we set self-pruning ratio to 0.7.

% \noindent\textbf{Baseline methods.} We selected several open-source implementations for comparison with our HCPM method, encompassing both conventional and recent research approaches. For detector-based methods, we chose SuperGlue~\cite{sarlin2020superglue}, SGMNet~\cite{chen2021learning}, and LightGlue~\cite{lindenberger2023lightglue}, all of which are combined with the SuperPoint~\cite{Detone2018superpoint} extractor. For detector-free methods, we selected LoFTR~\cite{sun2021loftr}, the pioneering work in detector-free methods, as well as QuadTree~\cite{tang2022quadtree} and AspanFormer~\cite{chen2022aspanformer}, which redesigned the LoFTR transformer module. Additionally, we included recent works addressing scale variation, such as AdaMatcher~\cite{huang2022adaptive}, along with other recent research contributions. Furthermore, as the HPatches datasets contain more informative candidates with most pixels being static and suitable for matching, we set the self-pruning ratio to 0.7.

\noindent\textbf{Results.} As shown in Table~\ref{tab:HPatches_result}, we observe that detector-based methods generally exhibit shorter runtimes compared to detector-free methods. However, our HCPM method, which is based on the detector-free approach, outperforms all the detector-based methods in terms of performance metrics while maintaining a runtime nearly identical to SuperGlue~\cite{sarlin2020superglue}. Although our method's accuracy is slightly lower compared to some other detector-free methods, the improvement in speed is more significant. This trade-off between accuracy and speed underscores the efficiency of our approach, positioning it as a competitive alternative for homography estimation tasks.

\begin{table}[htp]
	\centering
	\caption{\textbf{Evaluation on MegaDepth~\cite{li2018megadepth} for outdoor relative position estimation.}}
	\label{tab:MegaDepth_result}
	% %\vspace{-3mm}
	\scalebox{0.8}{
		\begin{tabular}{l c c c c c c}
			\hline
            \multicolumn{1}{c}{\multirow{2}{*}{Method}} &\multicolumn{3}{c}{Pose estimation AUC} & \multicolumn{3}{c}{Time cost(ms/per)}\\
			\cline{2-7}
                    &{@$5^\circ$}  &{@$10^\circ$}  &{@$20^\circ$} & module & metric & total\\
			\hline
            SuprGlue~\cite{sarlin2020superglue}   & 38.4 & 56.6 & 72.1 & 85.9 & 333.8 & 419.7 \\
        SGMNet~\cite{chen2021learning}     & 31.9 & 50.3  & 66.6 & 167.3 & 329.6 & 496.9 \\
        LightGlue~\cite{lindenberger2023lightglue}  & 35.7 & 54.7& 70.8 & 59.7 & 363.7 & 423.4 \\
        \cornerarrow 1600  & 50.2 & 67.7 & 80.4 & 128.7 & 241.0 & 369.7 \\
        \midrule 
        LoFTR~\cite{sun2021loftr}       & 52.8 & 69.2  & 81.2 & 181.0 & 122.9 & 303.9 \\
        Quadtree~\cite{tang2022quadtree}    & \textbf{53.9} & \textbf{70.4} & \textbf{82.1} & 265.0 & \textbf{100.9} & 365.9 \\
        AspanFormer~\cite{chen2022aspanformer} & 53.7 & 69.9  & 81.7 &  222.0 & 102.5 & 324.5 \\
        MatchFormer~\cite{wang2022matchformer} & 53.0 & 69.7 & 81.9  & 401.2 & 116.5 & 517.7 \\
        Adamatcher~\cite{huang2022adaptive}  & 52.6  & 69.6  & 81.8 & 396.2 & 101.9 & 498.1 \\
        HCPM        & 52.6 & 69.2 & 81.4 & 133.8  & 119.4  & 253.2 \\
        \cornerarrow FP16  & 51.3 & 68.1 & 80.7 & \textbf{94.2} & 112.2& \textbf{205.4} \\
        \hline
		\end{tabular}
}
\vspace{-3mm}
\end{table}

\subsection{Relative Pose Estimation}
For relative pose estimation, the majority of research employs the MegaDepth dataset ~\cite{Li2018} to showcase the effectiveness of their methods in outdoor scenes. In accordance with ~\cite{sun2021loftr}, we utilize the MegaDepth-1500 for testing, which comprises 1500 pairs from two independent scenes, demonstrating viewpoint and scale variations, as well as illumination changes.

\noindent\textbf{Evaluation protocol.} We use LoFTR~\cite{sun2021loftr} as the baseline since previous evaluations of various methods did not consistently use the same resolution, RANSAC method, or threshold. Therefore, we standardized all methods by evaluating them under the LoFTR framework. We use images with a resolution of 840x840, where the long side is 840 and the short side is padded to match. In our evaluation protocol, the relative poses are recovered from the essential matrix, estimated from feature matching with OpenCV RANSAC method with a threshold of 0.5. Following \cite{sarlin2020superglue}, we report the \textbf{AUC} of the pose error under thresholds ($5^{\circ}, 10^{\circ}, 20^{\circ}$), where the pose error is set as the maximum angular error of relative rotation and translation.

\noindent\textbf{Baseline methods.} Following the Hpatches evaluation baseline methods, we selected the same methods for evaluation on the Megadepth dataset. We compare our HCPM method with three detector-based methods, including SuperGlue~\cite{sarlin2020superglue}, SGMNet~\cite{chen2021learning}, and LightGlue~\cite{lindenberger2023lightglue}, as well as five detector-free methods, such as LoFTR~\cite{sun2021loftr}, QuadTree~\cite{tang2022quadtree}, AspanFormer~\cite{chen2022aspanformer}, and AdaMatcher~\cite{huang2022adaptive}. LightGlue~\cite{lindenberger2023lightglue} is a recently developed efficient method, which is faster than other methods. To ensure a fair comparison between detector-free and detector-based methods, we evaluate LightGlue using images with a long side resolution of 1600, as claimed in their paper, to keep the time cost approximately equal while comparing matching accuracy.
% This comprehensive comparison allows us to assess the performance of our HCPM method against various state-of-the-art approaches and demonstrate its effectiveness and efficiency in outdoor relative pose estimation task.

\noindent\textbf{Results.} Table.~\ref{tab:MegaDepth_result} demonstrates that detector-based methods typically exhibit shorter runtimes compared to detector-free methods. However, when calculating the pose based on existing matches, detector-based methods take significantly longer than their detector-free counterparts. The RANSAC method we use estimates the pose through multiple iterations by selecting points and terminating iterations based on the inlier ratio and error. This suggests that detector-free methods yield higher matching accuracy, leading to fewer iterations. Our method, based on LoFTR~\cite{sun2021loftr}, significantly reduces computation time while maintaining comparable accuracy. In comparison with the state-of-the-art LightGlue~\cite{lindenberger2023lightglue} method, our approach achieves better performance metrics with a runtime nearly identical. Moreover, by utilizing FP16 precision, our method achieves even faster processing speeds. Although there is a slight decrease in accuracy, our method demonstrates a more substantial improvement in speed, highlighting its efficiency and effectiveness in outdoor relative pose estimation tasks.

\subsection{Ablation Study}
We report the ablation results in Table.~\ref{tab:ablation}. The full HCPM method, which employs both self-pruning and implicit-pruning, achieves competitive performance compared to LoFTR~\cite{sun2021loftr} and LightGlue~\cite{sun2021loftr}. We further investigate the impact of each pruning strategy by removing self-pruning (a), interactive-pruning (b), and replacing implicit-pruning with direct interactive-pruning (c). The results indicate that removing self-pruning (a) leads to a slight drop in performance and an increase in computation time to 178.7 ms/per. Similarly, removing interactive-pruning (b) also results in a decrease in performance, although the computation time is slightly reduced to 140.7 ms/per. Lastly, when replacing implicit-pruning with direct interactive-pruning (c), the performance slightly declines, but the computation time is further reduced to 122.8 ms/per.

\begin{table}[htp]
	\centering
	\caption{\textbf{Ablation study on MegaDepth-1500~\cite{li2018megadepth}}.}
	\label{tab:ablation}
    \resizebox{0.45\textwidth}{!}{
    \begin{tabular}{l c c c c}
        \hline
        \multicolumn{1}{c}{\multirow{2}{*}{Method}} &\multicolumn{3}{c}{Pose estimation AUC} & \multicolumn{1}{c}{\multirow{2}{*}{ms/per}}\\
        \cline{2-4}
                &{@$5^\circ$}  &{@$10^\circ$}  &{@$20^\circ$} & \\
        \midrule
        LoFTR~\cite{sun2021loftr}~\tiny{CVPR'21} & 52.8 & 69.2 & 81.2 & 181.0\\
        LightGlue~\cite{sun2021loftr}~\tiny{ICCV'23} & 50.2 & 67.7 & 80.4 & 128.7\\
        \midrule[0.2pt]
        \b{HCPM} & 52.6 & 69.2 & 81.4 & 133.8  \\
        \cornerarrow a) w/o self-pruning & 51.9 & 68.4 & 80.5 & 178.7   \\
        \cornerarrow b) w/o interactive-pruning &51.0&67.8&80.3&140.7   \\
        \cornerarrow c) w direct interactive-pruning & 51.4 & 68.2 & 80.7 & 122.8   \\
        \bottomrule
    \end{tabular}}
\vspace{-3mm}
\end{table}

\subsection{Further Studies}

% 比较不同self-pruning ratio对性能和速度的影响，插入一个图即可
\noindent\textbf{Self-pruning ratio.} The self-pruning ratio significantly impacts the performance and efficiency of image matching. As shown in the Figure.~\ref{fig:self-pruning-ratio} below, selecting an appropriate self-pruning ratio is crucial for balancing performance and efficiency in image matching. As the first pruning stage, a smaller self-pruning ratio retains fewer candidates for the next module, leading to fewer matching candidate results. From the table, we can observe that a self-pruning ratio of 0.5 provides a good balance between performance and efficiency. As the ratio increases beyond 0.5, the performance improvement becomes marginal, while the time consumption continues to grow.
\begin{figure}[htp]
    \centering
    \includegraphics[width=0.45\textwidth]{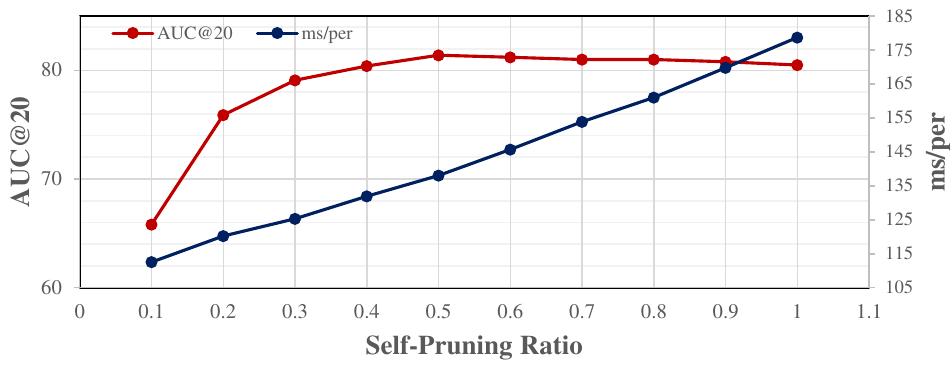}
    \caption{\textbf{Self-Pruning Ratio Analysis.}}
    \label{fig:self-pruning-ratio}
\vspace{-3mm}
\end{figure}

% ineractive-pruning不同设计的影响，是否直接丢掉还是sequeeze还是怎么处理之前
\noindent\textbf{Interactive-pruning.}
We have compared various interactive-pruning methods, including implicit pruning and direct pruning, as illustrated in Table.~\ref{tab:ablation}. In the HCPM model, we adopt the implicit interactive-pruning approach. Throughout the pipeline, we apply the interactive-pruning process in transformer blocks, which run $N$ times. We can choose to supervise either only the last block's selection result or all blocks' selection results. In Table.~\ref{tab:ablation_interactive}, the symbol \textbf{S} set to True indicates the use of the last block's selection results for supervision, and vice versa. Furthermore, after the interactive-pruning process, we can eliminate some unselected candidates. In Table.~\ref{tab:ablation_interactive}, the symbol \textbf{D} signifies the discarding of unselected candidates. 
Supervising with full DISC results leads to a minor decrease, primarily due to the early stages not being sufficiently interactive when supervising all DISC results. With direct discarding, we observe that some co-visible areas may be inaccurate at the pixel level, owing to depth loss and pose error of the ground truth. As OETR~\cite{chen2021learning} also suggests, this can lead to some incorrect supervision. Overlap estimation consistently overlooks certain slim candidates, such as those in pointy roofs. Consequently, directly discarding candidates might cause the loss of accurate candidates.

\begin{table}[htp]
	\centering
	\caption{\textbf{Interactive-pruning supervision and post-process}.}
	\label{tab:ablation_interactive}
    \resizebox{0.35\textwidth}{!}{
    \begin{tabular}{l c c c c c c}
        \hline
        \multicolumn{1}{c}{\multirow{2}{*}{Method}} &\multicolumn{1}{c}{\multirow{2}{*}{\textbf{S}}} &\multicolumn{1}{c}{\multirow{2}{*}{\textbf{D}}} &\multicolumn{3}{c}{Pose estimation AUC} & \multicolumn{1}{c}{\multirow{2}{*}{ms/per}}\\
        \cline{4-6}
               & & &{@$5^\circ$}  &{@$10^\circ$}  &{@$20^\circ$} & \\
        \midrule
        % \multicolumn{1}{c}{\multirow{4}{*}{Direct}} &\checkmark & \checkmark & \\
        %  &\checkmark &  & 51.4&68.2&80.7&122.8 \\
        %  & & \checkmark & \\
        %  & &  & &  &  & \\ \midrule
        \multicolumn{1}{c}{\multirow{4}{*}{Implicit}} &\checkmark & \checkmark & 44.8 & 62.6 & 76.3 & 125.6\\
         &\checkmark &  &52.6 & 69.2 & 81.4 & 133.8 \\
         & & \checkmark & 42.7 & 61.2&75.8&124.7\\
         & &  &  51.0 & 57.8 & 80.4 & 134.5 \\
        \bottomrule
    \end{tabular}}
\vspace{-5mm}
\end{table}

\section{Conclusion}
In this paper, we present HCPM, an innovative approach to local feature matching addressing the accuracy-efficiency trade-off in detector-free methods. HCPM employs a hierarchical pruning process, consisting of an initial self-pruning stage and an interactive-pruning phase, which aggregates information and removes uninformative candidates. By utilizing co-visible area supervision for a differentiable selection strategy, HCPM enhances matching descriptor performance and reduces redundancy. Drawing inspiration from token pruning techniques applied in segmentation and detection tasks, HCPM adapts these techniques for image matching tasks, retaining the dense benefits of detector-free methods while minimizing computational complexity. Additionally, we propose Gumbel-Softmax learned masks to automate the selection process, improving the method's overall efficiency and effectiveness. Our proposed HCPM method is poised to offer valuable insights to the feature matching community and drive further advancements in efficient local feature matching techniques.
%------------------------------------------------------------------------
%%%%%%%%% REFERENCES
\clearpage
{\small
\bibliographystyle{ieee_fullname}
\bibliography{egbib}
}
\end{document}

\title{Supplementary Material $for$ \\HCPM: Hierarchical Candidates Pruning for Efficient Detector-Free Matching}

\author{First Author\\
Institution1\\
Institution1 address\\
{\tt\small firstauthor@i1.org}
\and
Second Author\\
Institution2\\
First line of institution2 address\\
{\tt\small secondauthor@i2.org}
}
\maketitle

\section{Implementation Details}
\subsection{Interactive-Pruning Supervision}
To supervise the interactive-selection result, we leverage the co-visible area for guidance. However, employing the point-to-point co-visible area directly may result in inconsistencies and depth separation, posing challenges for learning. Aiming to guide the matching process without sacrificing crucial information, we devise an alternative approach. Rather than relying on the point-to-point co-visible area, we employ a co-visible bounding box that encompasses all valid depth pixels within the shared viewing region. This concept, inspired by OETR~\cite{chen2022guide}, involves using a co-visible box and subsequently evaluating the depth validation within that box. Consequently, when the depth in the co-visible box exceeds zero, it is deemed valid. Utilizing this method allows for better information preservation and enhances the performance of our interactive-selection process.

\begin{figure}[htp]
\centering
\resizebox{.48\textwidth}{!}{
    \begin{tabular}{@{}cc@{}}
        \includegraphics[]{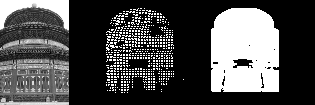} &
        \includegraphics[]{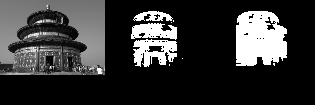} \\
        \includegraphics[]{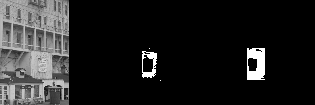} &
        \includegraphics[]{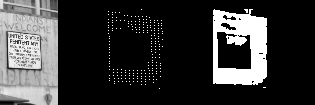} \\ 
        \includegraphics[]{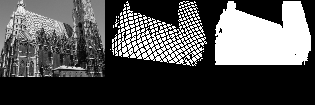} &
        \includegraphics[]{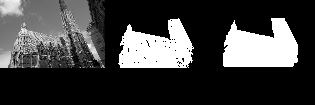} \\ 
        \includegraphics[]{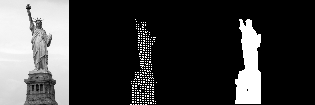} &
        \includegraphics[]{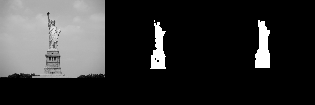} \\ 
        \includegraphics[]{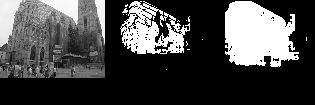} &
        \includegraphics[]{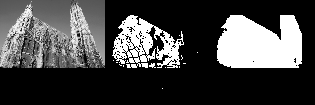} \\
    \end{tabular}
}
\caption{\textbf{Interactive-pruning supervision visualization.} Visualizations for five matching pairs, each representing a different scene. In each row, two images with varying viewpoints or scales are displayed. For every image, the first mask represents a one-to-one correspondence mask, while the second mask illustrates the valid depth region within the bounding box. These visualizations demonstrate the effectiveness of our interactive-pruning supervision approach across diverse scenes and conditions.}
\label{fig:co_mask}
\end{figure}

% \subsection{Scale-alignment int Fine-level Matching}
% As demonstrated in ASTR~\cite{yu2023adaptive} and AdaMatcher~\cite{huang2022adaptive}, the co-visible area is not directly used to select matching candidates. Instead, the ratio between the co-visible areas of the two images is indirectly employed to align the feature map at a fine level. In our approach, we utilize the DICS to obtain the co-visible candidates. Although it is not precise enough to guide the selection of matching candidates, it can still be used as a rough scale ratio to align the feature map.

\section{More Experiments}
\subsection{Computational Cost Comparision}
In this section, we compare our method, HCPM, with LoFTR and LightGlue in terms of FLOPS, parameters, and model size. The results are shown in Table~\ref{tab:computation}. We use the `thop` library to compute the parameter size and FLOPS, and the `nvidia-smi` command in the pipeline to acquire the maximum memory usage.

\begin{table}[h]
\centering
\caption{Computational complexity of different methods.}
\label{tab:computation}
\begin{tabular}{@{}c|ccc@{}}
\toprule
\multicolumn{1}{c}{\multirow{1}{*}{Method}} & \multicolumn{1}{c}{Param(MB)} & \multicolumn{1}{c}{Flops(G)} & \multicolumn{1}{c}{Memory(G)} \\ 
\midrule
LoFTR  & 11.17 & 799.90 & 5.77 \\  
SP+LightGlue & 10.12 & 141.66 & 2.81 \\
HCPM & 11.24 & 737.23 & 3.82 \\
\bottomrule
\end{tabular}
\end{table}

As shown in the table, our method, HCPM, has a similar number of parameters as LoFTR, but with significantly fewer FLOPS and lower memory usage. This demonstrates the efficiency of our method compared to the other approaches.

\subsection{Results on YFCC100M}\label{result_yfcc}
We further evaluate the performance of our method, HCPM, on the YFCC100M dataset~\cite{thomee2016yfcc100m}, comparing it with several baseline methods. All models are trained on the MegaDepth~\cite{Li2018} dataset, and the test set is derived from four selected landmark sequences, with each sequence sampling 1000 image pairs. The images are resized to 480 $\times$ 640, and we use the same test pairs (a total of 4000 pairs) and evaluation metrics as in previous works~\cite{sarlin2020superglue, oanet2019iccv} to ensure a fair comparison.

The accuracy of pose estimation is measured by AUC under error thresholds ($5^\circ$, $10^\circ$, and $20^\circ$). As shown in Table~\ref{tab:yfcc100m}, our method consistently performs faster than the corresponding baseline methods while maintaining nearly the same accuracy. This demonstrates the effectiveness of HCPM in relative pose estimation on the YFCC100M dataset.

\begin{table}[htp]
\caption{Evaluation on YFCC100M~\cite{thomee2016yfcc100m} for relative pose estimation.}
\label{tab:yfcc100m}
\resizebox{.46\textwidth}{!}{
\begin{tabular}{lllll}
\toprule
\multicolumn{1}{l}{\multirow{2}{*}{\textbf{Method}}} & \multicolumn{3}{c}{\textbf{Pose est. AUC}}   &\multicolumn{1}{l}{\multirow{2}{*}{\textbf{ms/per}}}     \\\cline{2-4}
            & @5px                   & @10px                   & @20px                                    \\\midrule
% SuperGlue~\cite{sarlin2020superglue}~\tiny{CVPR'19}   & \multicolumn{1}{r}{27.87} & \multicolumn{1}{r}{47.44} & \multicolumn{1}{r}{65.87} & \multicolumn{1}{r}{59.16}    \\
% LightGlue~\cite{lindenberger2023lightglue}~\tiny{ICCV'23}   & \multicolumn{1}{r}{30.57} & \multicolumn{1}{r}{51.03} & \multicolumn{1}{r}{68.81} & \multicolumn{1}{r}{48.07}     \\
% \midrule
LoFTR~\cite{sun2021loftr}~\tiny{CVPR'21}       & \multicolumn{1}{r}{43.06} & \multicolumn{1}{r}{62.20} & \multicolumn{1}{r}{77.25} & \multicolumn{1}{r}{71.16} \\
QuadTree~\cite{tang2022quadtree}~\tiny{ICLR'22}   & \multicolumn{1}{r}{39.38} & \multicolumn{1}{r}{58.37} & \multicolumn{1}{r}{73.50}    & \multicolumn{1}{r}{111.93}  \\
AspanFormer~\cite{chen2022aspanformer}~\tiny{ECCV'22} & \multicolumn{1}{r}{42.47} & \multicolumn{1}{r}{61.58} & \multicolumn{1}{r}{76.60} & \multicolumn{1}{r}{120.23} \\
AdaMatcher~\cite{huang2022adaptive}~\tiny{CVPR'23}        & \multicolumn{1}{r}{44.11} & \multicolumn{1}{r}{63.08} & \multicolumn{1}{r}{77.64} & \multicolumn{1}{r}{233.44} \\
HCPM        & \multicolumn{1}{r}{42.64} & \multicolumn{1}{r}{61.78} & \multicolumn{1}{r}{76.76} & \multicolumn{1}{r}{67.89} \\\bottomrule  
\end{tabular}}
\end{table}

\section{Qualitative Results}
\begin{figure*}[htp]
\centering
\resizebox{.95\textwidth}{!}{
    \begin{tabular}{@{}ccccc@{}}
         \multicolumn{1}{c}{HCPM} & \multicolumn{1}{c}{Adamatcher} & \multicolumn{1}{c}{AspanFormer} & \multicolumn{1}{c}{LoFTR} & \multicolumn{1}{c}{QuadTree}\\% \multicolumn{1}{c}{SuperGlue}& \multicolumn{1}{c}{LightGlue}\\
        \includegraphics[]{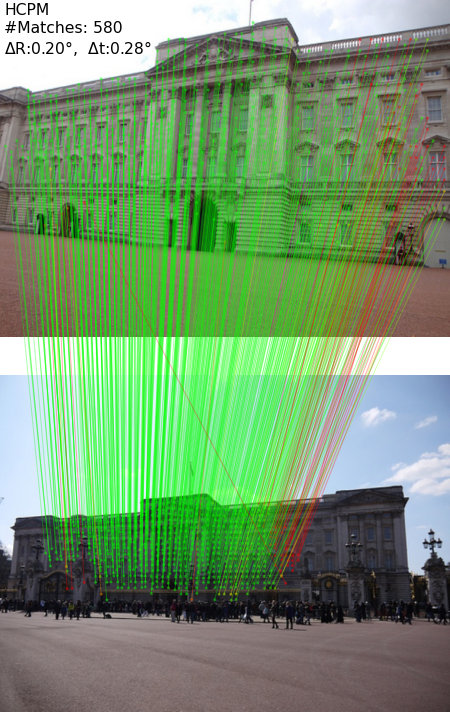} &
        \includegraphics[]{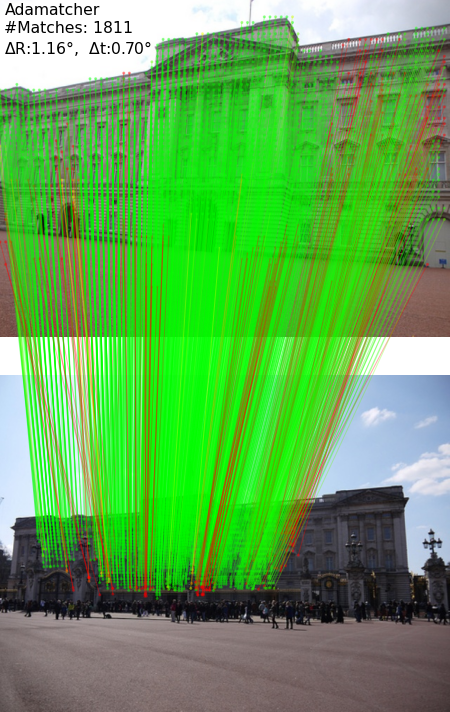} &
        \includegraphics[]{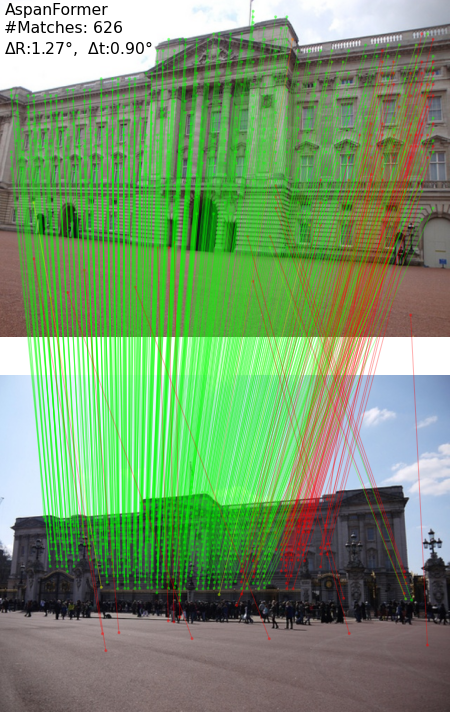} &
        \includegraphics[]{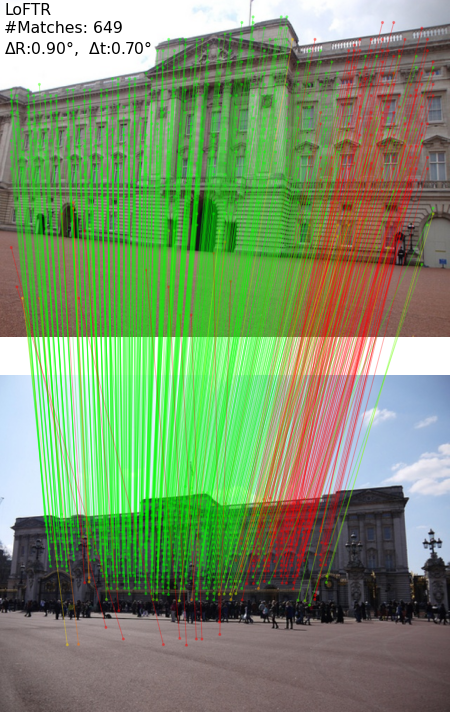} &
        \includegraphics[]{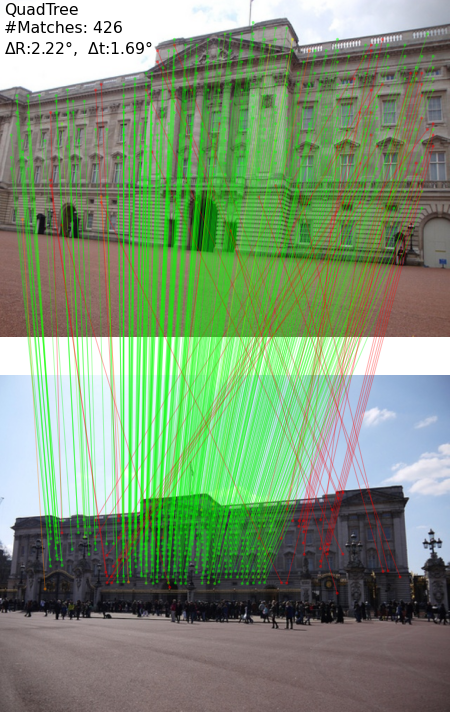} \\ 
        \includegraphics[]{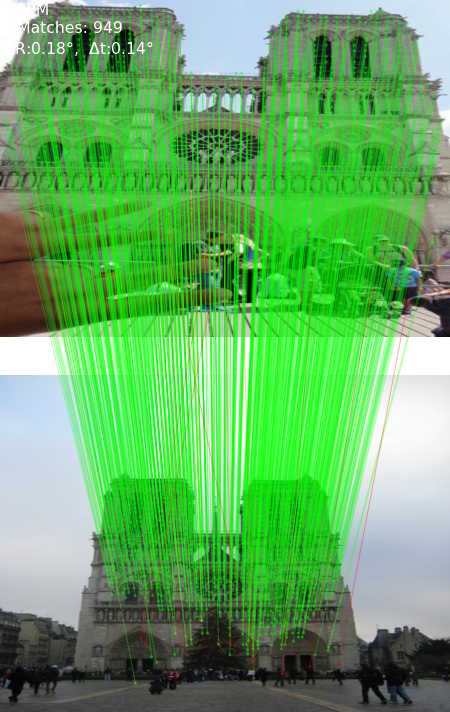} &
        \includegraphics[]{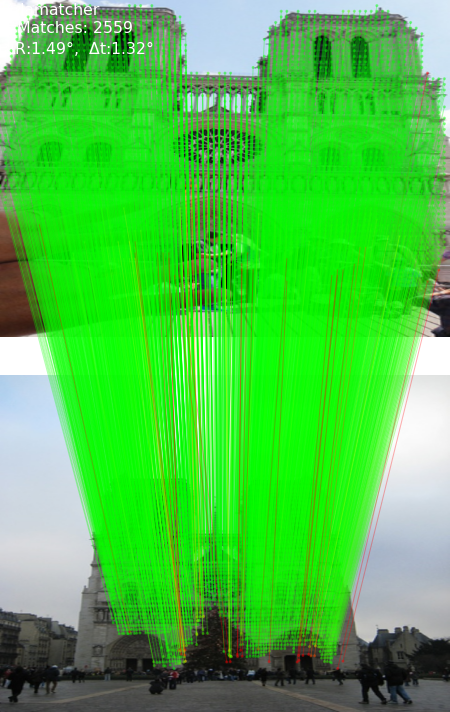} &
        \includegraphics[]{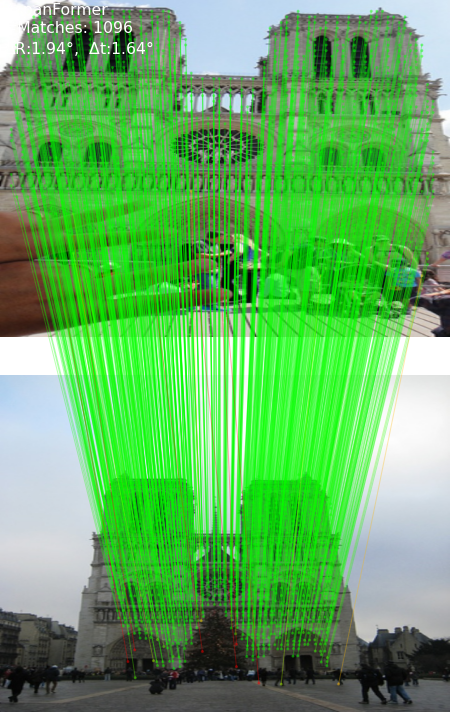} &
        \includegraphics[]{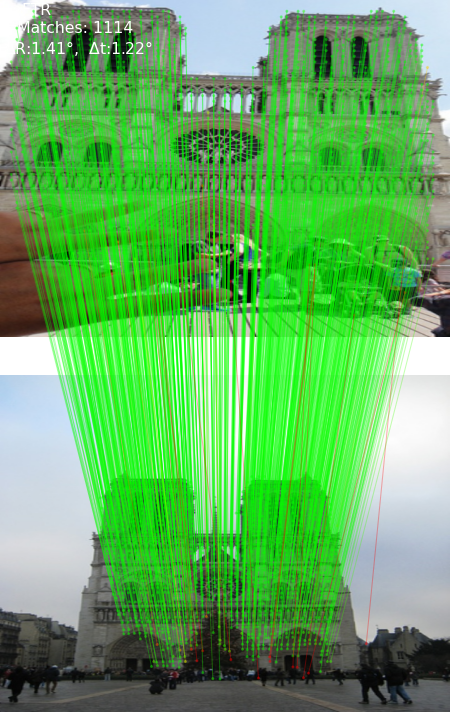} &
        \includegraphics[]{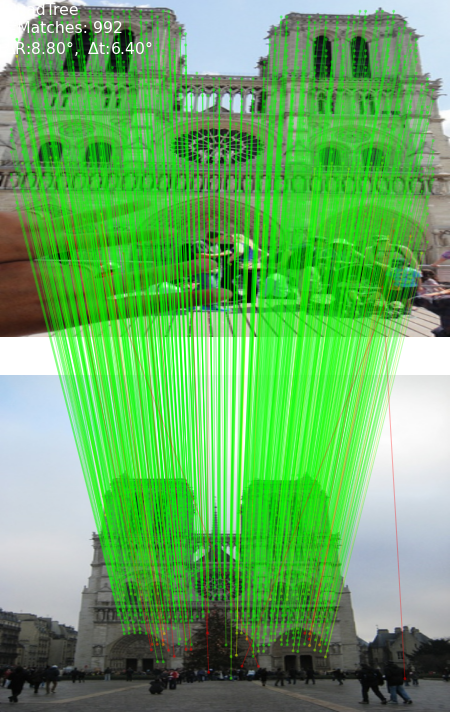} \\
        \includegraphics[]{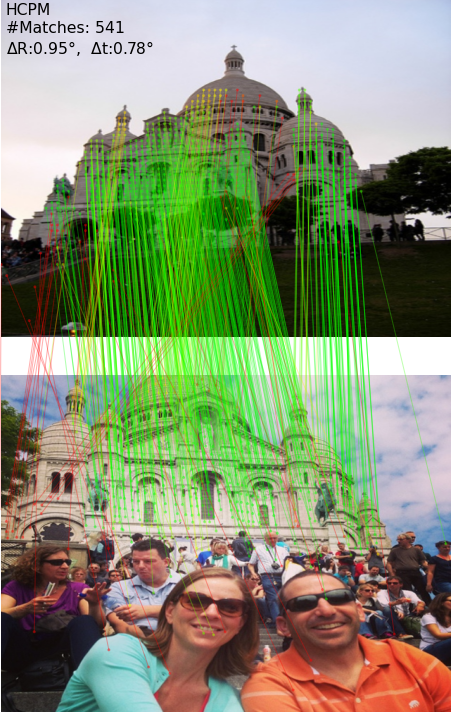} &
        \includegraphics[]{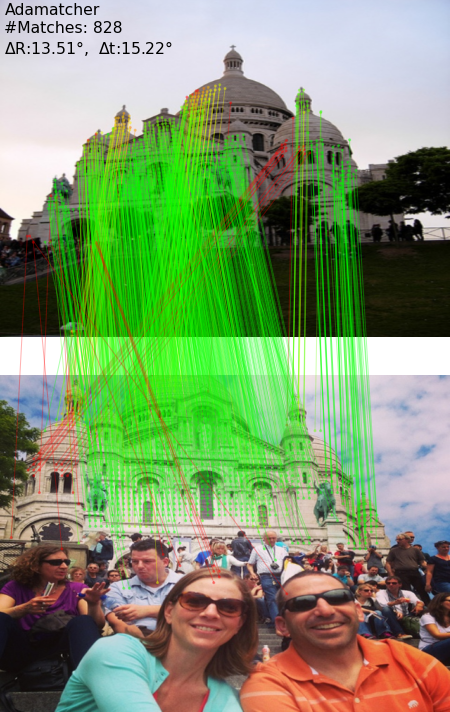} &
        \includegraphics[]{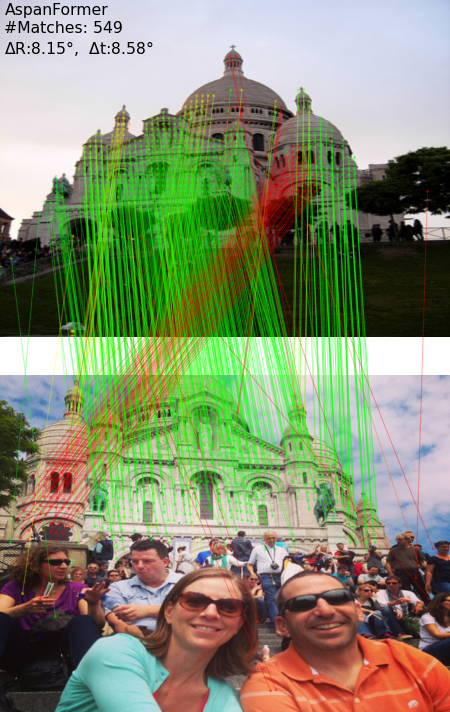} &
        \includegraphics[]{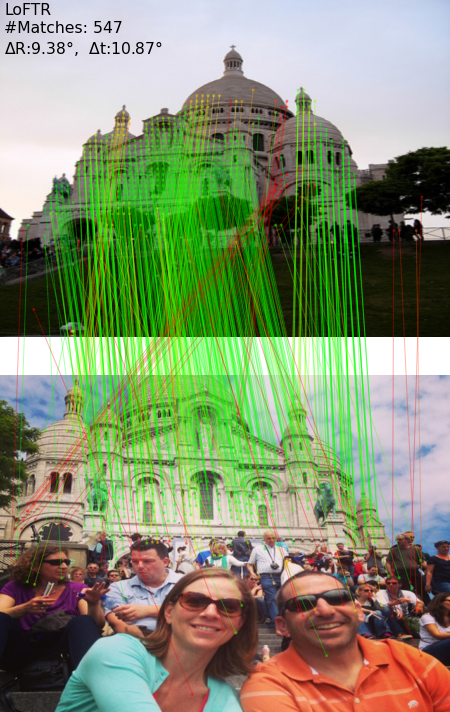} &
        \includegraphics[]{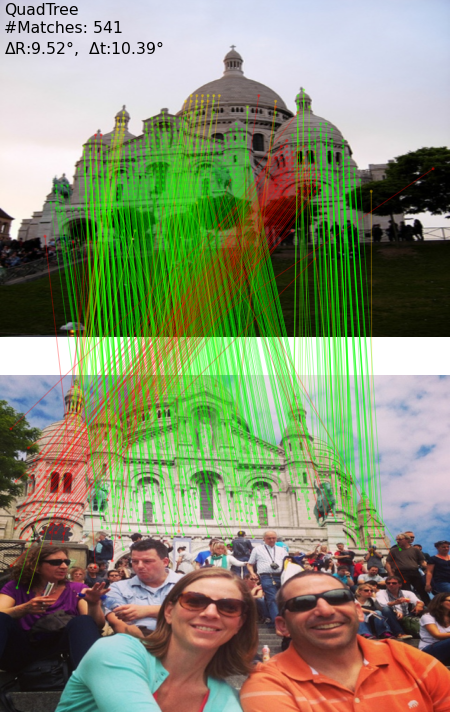} \\
        \includegraphics[]{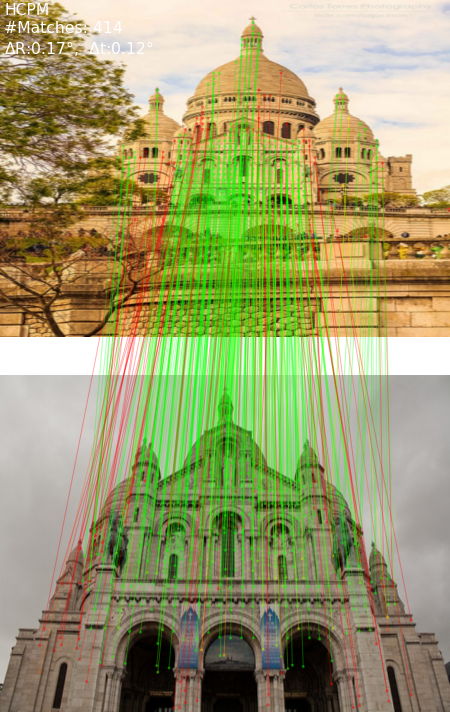} &
        \includegraphics[]{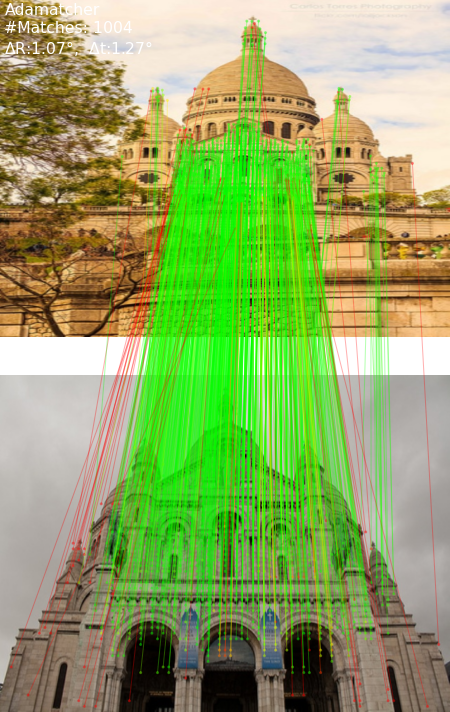} &
        \includegraphics[]{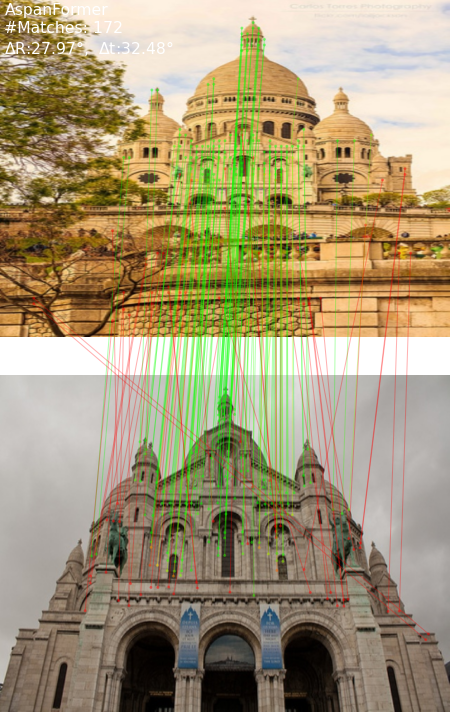} &
        \includegraphics[]{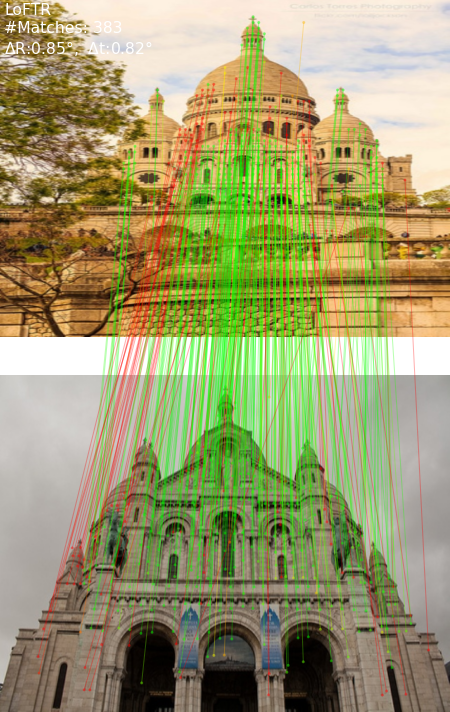} &
        \includegraphics[]{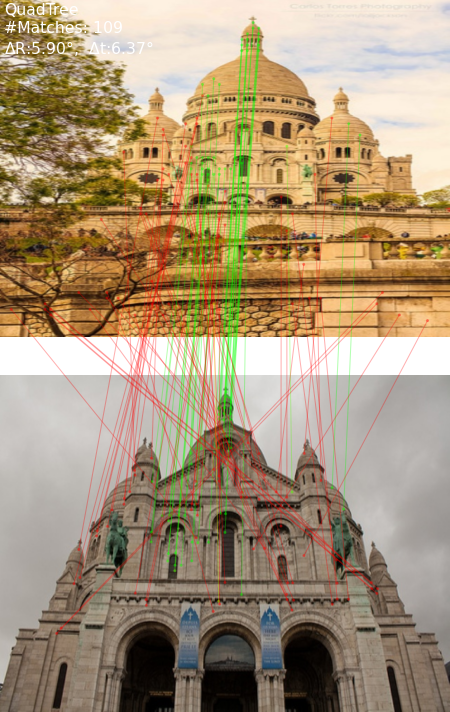} \\
    \end{tabular}
}
\caption{\textbf{YFCC100M Visualization.}  Green indicates that epipolar error in normalized image coordinates is less than $1\times 10^{-4}$, while red indicates that it is exceeded.}
\label{fig:yfcc}
\end{figure*}

We present qualitative comparisons of HCPM and baseline methods on the YFCC100M~\cite{thomee2016yfcc100m} dataset. In Fig.\ref{fig:yfcc}, we display inlier and outlier matches using different projection thresholds to compare the matching accuracy of various methods on the YFCC100M dataset. The columns in the figure represent, from left to right, HCPM, AdaMatcher~\cite{huang2022adaptive}, AspanFormer~\cite{chen2022aspanformer}, LoFTR~\cite{sun2021loftr}, and QuadTree~\cite{tang2022quadtree}, respectively.

%%%%%%%%% REFERENCES
\clearpage
{\small
\bibliographystyle{ieee_fullname}
\bibliography{egbib}
}